\documentclass[11pt]{article}
\title{On the Sequence of State Configurations in the Garden of Eden}
\author{Yukihiro KAMADA \thanks{Follows information processing education at Tohoku Fukushi University. D.Eng} \and Kiyonori MIYASAKI \thanks{Professor Emeritus at Tohoku Institute Technology}}
\date{}
\usepackage{graphicx}

\begin{document}
\maketitle

\begin{abstract}
Autonomous threshold element circuit networks are used to investigate the structure of neural networks. With these circuits, as the transition functions are threshold functions, it is necessary to consider the existence of sequences of state configurations that cannot be transitioned. In this study, we focus on all logical functions of four or fewer variables, and we discuss the periodic sequences and transient series that transition from all sequences of state configurations. Furthermore, by using the sequences of state configurations in the Garden of Eden, we show that it is easy to obtain functions that determine the operation of circuit networks.
\end{abstract}

\section{Introduction}
Autonomous threshold element circuit networks are gaining increased attention as a logical model for neural networks, and corresponding studies regarding the operation of such networks are therefore also increasing [1]. Furthermore, numerous studies have focused on the structure of the sequences of state configurations to investigate the operation of circuit networks [2-8]. Majority of these studies performed logical analysis in relation to structural elements in circuit networks, and the clarified nature of the elements often only has local validity in relation to the circuit networks. This has occurred because most of the global sequences of state configurations in a circuit network were emergently created by the transition functions. As an example of such logical analysis, one proposed method discovered common areas between the characteristics of the sequences of state configurations and the nature of the threshold function. It did so by viewing the autonomous circuit network as a single threshold element; however, these discussions often returned to the structural analysis of a one-variable function. Since one-variable functions can be realized by single threshold functions, the nature of one-variable functions can be locally discovered in relation to circuit networks.

A number of studies also used the nature of elements when generating circuit networks by viewing the entire circuit network as one element [9-19]. A circuit network has been reported that can realize a parity function by concentrating on the connection rather than the element. The connection from the output layer to the input layer represented the dimensional extension of the input layer element. As a result, it was
possible to embed the function of the hidden layer element in a high-dimension input layer; however, this report effectively states that gthe greater also serves the lesser.h Therefore, this study did not seem relevant to our research.

In our study, we generated sequences of state configurations in relation to all logical functions consisting of two to four variables; this is accomplished by applying certain transition functions. Here, we show the maximum cycles in the periodic sequences that emerged during the generation process and the total number of sequences of state configurations in the Garden of Eden. Furthermore, we show that it was possible to create specific sequences of state configurations by using the obtained results as genetic algorithm parameters.

\section{Display of Autonomous Circuit Network Formulas}
An autonomous circuit network is a circuit network realizing an arbitrary logical function to which the concept of time has been added. This behavior is determined as follows:
\begin{equation}
x(z+1)=f(X_z)
\end{equation}
Here, $x(z+1) \in \left\{ 0, 1 \right\}$ expresses the state of a circuit network at time $z+1$; $f$ is a disjunctive normal form for realizing $n$ variables, expressed as follows [1] in relation to
$n$-dimension input vector $X=(x_1, x_2, \ldots , x_n)$:
\begin{equation}
\begin{array}{lll}
f(X) & = & x_1 x_2 \cdots x_n f_{1 2 \cdots n} \vee \bar{x}_1 x_2 \cdots x_n f_{\bar{1} 2 \cdots n} \vee x_1 \bar{x}_2 \cdots x_n f_{1 \bar{2} \cdots n} \\ 
&  & \qquad   \vee \cdots \vee \bar{x}_1 \bar{x}_2 \cdots \bar{x}_n f_{\bar{1} \bar{2} \cdots \bar{n}} 
\end{array} 
\end{equation}
$X_z$ is an $n$-dimensional column vector represented by the following equation, known as the column state at time $z$.
\begin{equation}
X_z = (x(z+1), x(z), \ldots , x(z-(n-1)))^t
\end{equation}
Here, $t$ expresses transposition.

Considering the transition from sequences of state configurations $X_z$ to $X_{z+1}$ , the transition process is expressed as $X_z \to  X_{z+1}$. From equations (1) and (2), $X_{z+1}$ is expressed as follows:
\begin{equation}
X_{z+1} = (f(X_z), x(z), x(z-1), \ldots , x(z-(n-2)))^t
\end{equation}

\section{Sequences of State Configurations and their Transitions}
In regards to sequences of state configurations that realize autonomous circuit networks, where $X_z = X_{z+r}$ exists, such sequences are referred to as $r$-periodic sequences. When the sequences of state configurations are created using one logical function and multiple $r$ values, the maximum value for $r$ is referred to as the maximum cycles in the periodic sequence. Sequences of state configurations that do not contain periodic sequences are referred to as transient series; furthermore, transient series without preceding sequences of state configurations are referred to as sequences of state configurations in the Garden of Eden.

Using three-variable function $f$ with the truth table shown in Table 1, the sequences of state configurations are generated. Figure 1 illustrates this transition process. The sequence of state configurations $(0,0,0)^t$ is the periodic sequence that transitions to itself, as are $(0,0,1)^t$, $(1,0,0)^t$, $(0,1,0)^t$, and the maximum cycles for these are three. All other sequences of state configurations are transient series, but since $(1,0,1)^t$, $(1,1,1)^t$ do not have preceding sequences of state configurations, these are sequences of state configurations in the Garden of Eden.

\begin{table}[htbp]
 \caption{Truth Table for Three-Variable Function $f$}
 \begin{center}
  \begin{tabular}{ccc|c}
     $x_1$  & $x_2$   & $x_3$   & $f$   \\
    \hline
    0   & 0   & 0   & 0   \\
    0   & 0   & 1   & 1   \\
    0   & 1   & 0   & 0   \\
    0   & 1   & 1   & 0   \\
    1   & 0   & 0   & 0   \\
    1   & 0   & 1   & 1   \\
    1   & 1   & 0   & 0   \\
    1   & 1   & 1   & 0   \\
  \end{tabular}
 \end{center}
\end{table}

\begin{figure}[htbp]
  \begin{center}
    \includegraphics[width=100mm,height=100mm]{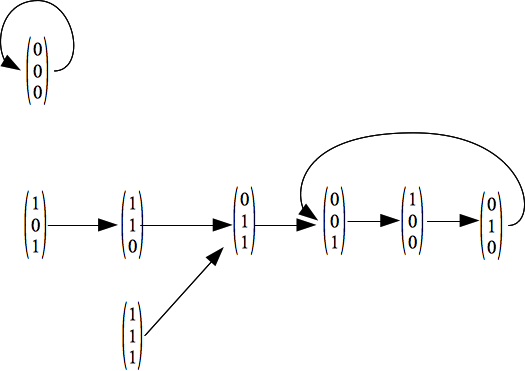}
  \end{center}
  \caption{State Transition Diagram for $f$}
\end{figure}

\section{Transitions in Logical Functions up to Four Variables}
Tables 2 and 3 show the maximum cycles and the total number of sequences of state configurations in the Garden of Eden for autonomous circuit networks using logical functions with two to four variables, respectively.

\begin{table}[htbp]
 \caption{Functions with Two to Four Variables that have Maximum $r$-Periodic Sequences}
 \begin{center}
  \begin{tabular}{c|ccr}
    \hline
    Maximum cycles   &    &  $n$  &    \\
    \cline{2-4}
    $r$   & 2   & 3   & 4 \hspace{2mm}    \\
    \hline
     1  &  7  & 60   & 7936   \\
     2  &  4  & 55   & 6864   \\
     3  & 5   & 56   & 11548   \\
     4  & --   & 47   & 9921   \\
     5  &  --  &  16  & 11274   \\
     6  &  --  &   12 & 7068   \\
     7  &  --  &  10  & 4064   \\
     8  &  --  & --   & 3069   \\
     9  & --   & --   & 1792   \\
     10  &  --  & --   &  1088  \\
     11  &  --  &  --  & 448   \\
     12  &  --  & --   & 208   \\
     13  & --   & --   & 96   \\
     14  &  --  & --   & 80   \\
     15  &  --  & --   & 80   \\
    \hline
  \end{tabular}
 \end{center}
\end{table}

\begin{table}[htbp]
 \caption{Number of Functions with Two to Four Variables in Relation to the Total Number $d$ of Sequences of State Configurations in the Garden of Eden}
 \begin{center}
  \begin{tabular}{c|ccr}
    \hline
    No. of sequences  of state configurations   &    &  $n$  &    \\
    \cline{2-4}
    $d$   & 2   & 3   & 4 \hspace{2mm}    \\
    \hline
     0  &  4  & 16   & 256   \\
     1  &  8  & 64   & 2048   \\
     2  &  4  & 96   & 7168   \\
     3  & --   & 64   & 14336   \\
     4  & --   & 16   & 17920   \\
     5  &  --  &  --  & 14336   \\
     6  &  --  &  --  & 7168   \\
     7  &  --  &  --  & 2048   \\
     8  &  --  & --   & 256   \\
    \hline
  \end{tabular}
 \end{center}
\end{table}

As shown in Table 2, the number of functions tends to decrease as the number of variables and maximum cycles increase. The number of sequences of state configurations in the Garden of Eden is expressed as $d$. Furthermore, as shown in Table 3, the number of functions constructed as only periodic sequences ($d=0$) and the number of functions that satisfy $d=2$ are equal. In the case of three or four variables, the value of $d$ differs but the same relationship is established.

\section{Generation of Sequences of State Configurations using Genetic Algorithms}
When investigating the maximum number of cycles in the case of $n$-variable functions, the number of sequences of state configurations exponentially increase with the increase in $n$. In this section, we describe the use of genetic algorithms to derive Maximum cycles logical functions with the specified maximum number of periodic sequences. Furthermore, we show that by simultaneously specifying the total number of sequences of state configurations in the Garden of Eden when specifying the maximum number of cycles, it becomes easier to obtain the objective logical functions.

Genetics $g$ using $n$-variable functions $f$ are determined as follows in relation to $n$-dimensional vector
$X_i=(x_1, x_2, \ldots , x_n)$, $i \in \left\{0, 1, \ldots , 2^n-1 \right\}$:
\begin{equation}
g=f(X_i)
\end{equation}

In equation (5), from the fact that functions and genes can be seen in the same light, maximum cycles can be examined to determine the fitness of the genes. First, the genes are applied to an autonomous circuit network to generate the sequences of state configurations. In concrete terms, all input vectors transition as $X_z \to X_{z+2^n}$ as a sequence of state configurations of time $z$.

Next, in the transition process, sequences of state configurations are classified into periodic sequences and transient series. When a periodic sequence is obtained, the maximum cycles $r$ are found, and when a transient series is obtained, we investigate whether there are preceding sequences of state configurations; if not, we count the number $d$ of such sequences of state configurations. The specified maximum cycles and sequences of state configurations in the Garden of Eden are expressed as $r'$, $d'$ and $m$ is determined as an appropriately large value.

Therefore, $m - (r'-r)^2 + (d'-d)^2$ is referred to as the fitness of gene $g$; furthermore, the genetic algorithm uses uniform crossover or elite selection. For spontaneous mutations, part of the gene is inverted from $0\to 1 $ (or $1 \to 0)$ with a probability of 1\%. The genetic population is 1000 and the upper limit for the number of times they can be genetically processed is set to 1 million generations.

Next, the maximum cycles is set to 12 and the total number of sequences of state configurations in the Garden of Eden is set to nine. Autonomous circuit network parameters were sought using a genetic algorithm that satisfied these conditions with results are shown in Figure 2.

\begin{figure}[htbp]
  \begin{center}
    \includegraphics[width=100mm,height=100mm]{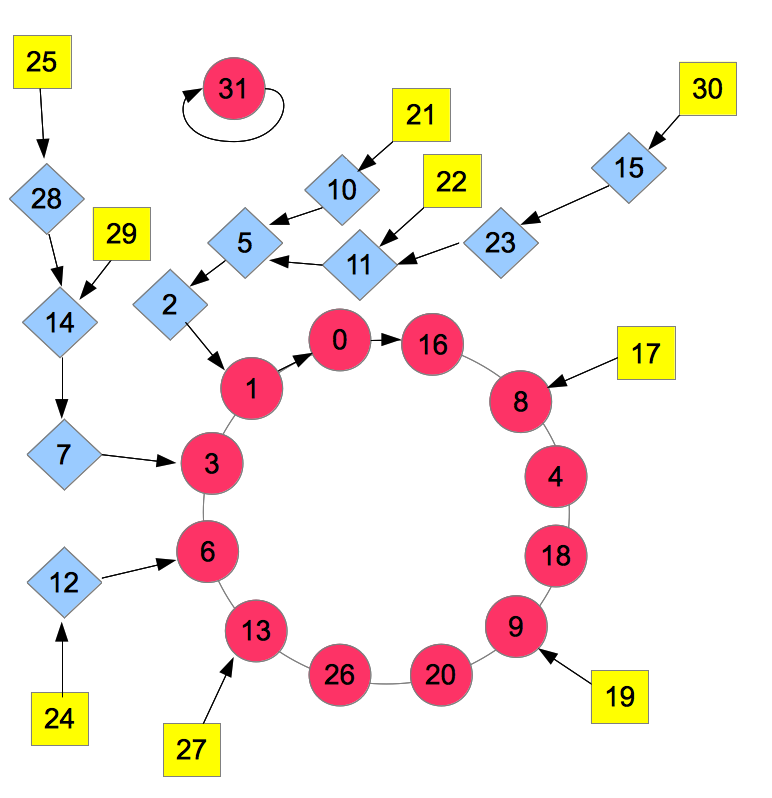}    
  \end{center}
  \caption{State Space Satisfying Maximum Cycles of 12 and Total Number of Sequences of State Configurations in the Garden of Eden of Nine}
\end{figure}

The red circles in figure 2 represent sequences of state configurations made from periodic sequences, whereas the blue rhombi indicate sequences of state configurations made from transient states, and the yellow squares show sequences of state configurations in the Garden of Eden. The numbers within the symbols display the sequences of state configurations in decimal form. The transient series are arranged in such a way that they are absorbed into the ringed sequences of state configurations. Here, 31 is the number of cycles to reach itself. In places where the total number of sequences of state configurations in the Garden of Eden is not used as a parameter in the genetic algorithm, the discovery of excellent genes is largely dependent on this initial value, and little increase in fitness was seen with an increase in the number of generations.

\section{Conclusions}
Although it is necessary to study the behavior of elements to analyze the structure of autonomous circuit networks, it is difficult to grasp the overall behavior of the circuit networks by only using this approach. In this study, we have clarified the maximum cycles and total number of sequences of state configurations in the Garden of Eden when transient conditions are applied in relation to all logical functions up to four variables. Furthermore, by investigating the behavior of high-dimension circuit networks, we showed that our results successfully functioned as genetic algorithm parameters. We conclude that this will provide useful clues to better understanding the behavior of logical models for neural networks. Our future work includes the application of theoretical studies to emergent behavior in autonomous circuit networks.

\end{document}